\documentclass[10pt,twocolumn,letterpaper]{article}

\usepackage{iccv}
\usepackage{times}
\usepackage{epsfig}
\usepackage{graphicx}
\usepackage{amsmath}
\usepackage{amssymb}
\usepackage[accsupp]{axessibility}
\usepackage{makecell}
\usepackage{multirow}
\usepackage{adjustbox}
\usepackage{textgreek}
\usepackage{booktabs}
\usepackage{bbding}

\iccvfinalcopy 

\def\iccvPaperID{3124} 

\ificcvfinal\fi

\begin{document}

\title{Aleth-NeRF: Low-light Condition View Synthesis with Concealing Fields}

\author{Ziteng Cui\textsuperscript{1,2}\thanks{This work was done during his internship at Shanghai Artificial Intelligence Laboratory. \textsuperscript{\Envelope} means corresponding author.}
    Lin Gu\textsuperscript{3,1} 
    Xiao Sun\textsuperscript{2}\textsuperscript{\Envelope}
    Xianzheng Ma\textsuperscript{4}
    Yu Qiao\textsuperscript{2}
    Tatsuya Harada\textsuperscript{1,3}\\
    \textsuperscript{1}The University of Tokyo \textsuperscript{2}Shanghai AI Laboratory \textsuperscript{3}RIKEN AIP \textsuperscript{4}The University of Oxford
    }

\maketitle
\ificcvfinal\thispagestyle{empty}\fi

\begin{abstract}
  Common capture low-light scenes are challenging for most computer vision techniques, including Neural Radiance Fields (NeRF). Vanilla NeRF is viewer-centred that simplifies the rendering process only as light emission from 3D locations in the viewing direction, thus failing to model the low-illumination induced darkness.  Inspired by emission theory of ancient Greek that visual perception is accomplished by rays casting from eyes, we make slight modifications on vanilla NeRF to train on multiple views of low-light scene, we can thus render out the well-lit scene in an \textbf{unsupervised} manner. We introduce a surrogate concept, Concealing Fields, that reduce the transport of light during the volume rendering stage. Specifically, our proposed method, \textbf{Aleth-NeRF}, directly learns from the dark image to understand volumetric object representation and concealing field under priors. By simply eliminating Concealing Fields, we can render a single or multi-view well-lit image(s)  and gain superior performance over other 2D low light enhancement  methods. Additionally, we collect the first paired \textbf{LO}w-light and normal-light \textbf{M}ulti-view (LOM) datasets for future research. Our code and dataset will be released soon. 
\end{abstract}

\section{Introduction}
\label{sec:intro}

\begin{figure}
    \centering
    \includegraphics[width=1.0\linewidth]{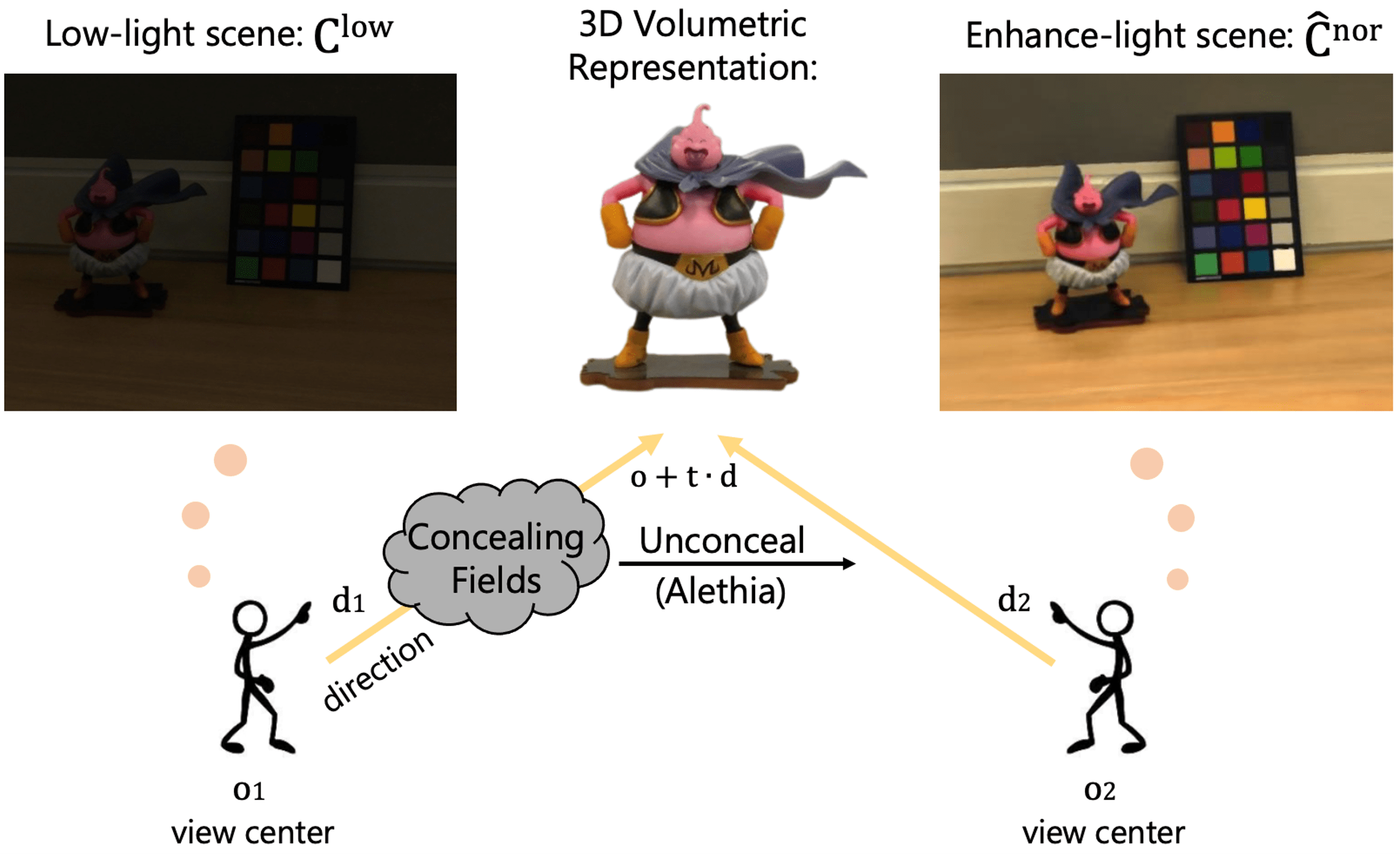}
    \caption{We assume objects are naturally visible. However, the Concealing Field attenuates the light in the viewing direction, making the left user see a low-light scene. Aleth-NeRF takes a low-light image as input and unsupervisly learns the distribution of the Concealing Field. Then, we unconceal (alethia) the Concealing field to render the enhanced image. This scene is taken from LOM dataset.}
    \vspace{-2mm}
    \label{pics:fig1}
\end{figure}

Neural Radiance Field (NeRF)~\cite{nerf} has been demonstrated to be effective in understanding 3D scenes from 2D images and generate novel views.
However, similar to most computer vision algorithms  such as semantic segmentation~\cite{WU_2021_CVPR,kerim2022Semantic}, object detection~\cite{HLAFace_2021_CVPR,Cui_2021_ICCV}, and \textit{etc.}, NeRF often fails in sub-optimal lighting scenes captured  under insufficient illumination or limited exposure time~\cite{raw_nerf}. 

\begin{figure}
    \centering
    \includegraphics[width=0.98\linewidth]{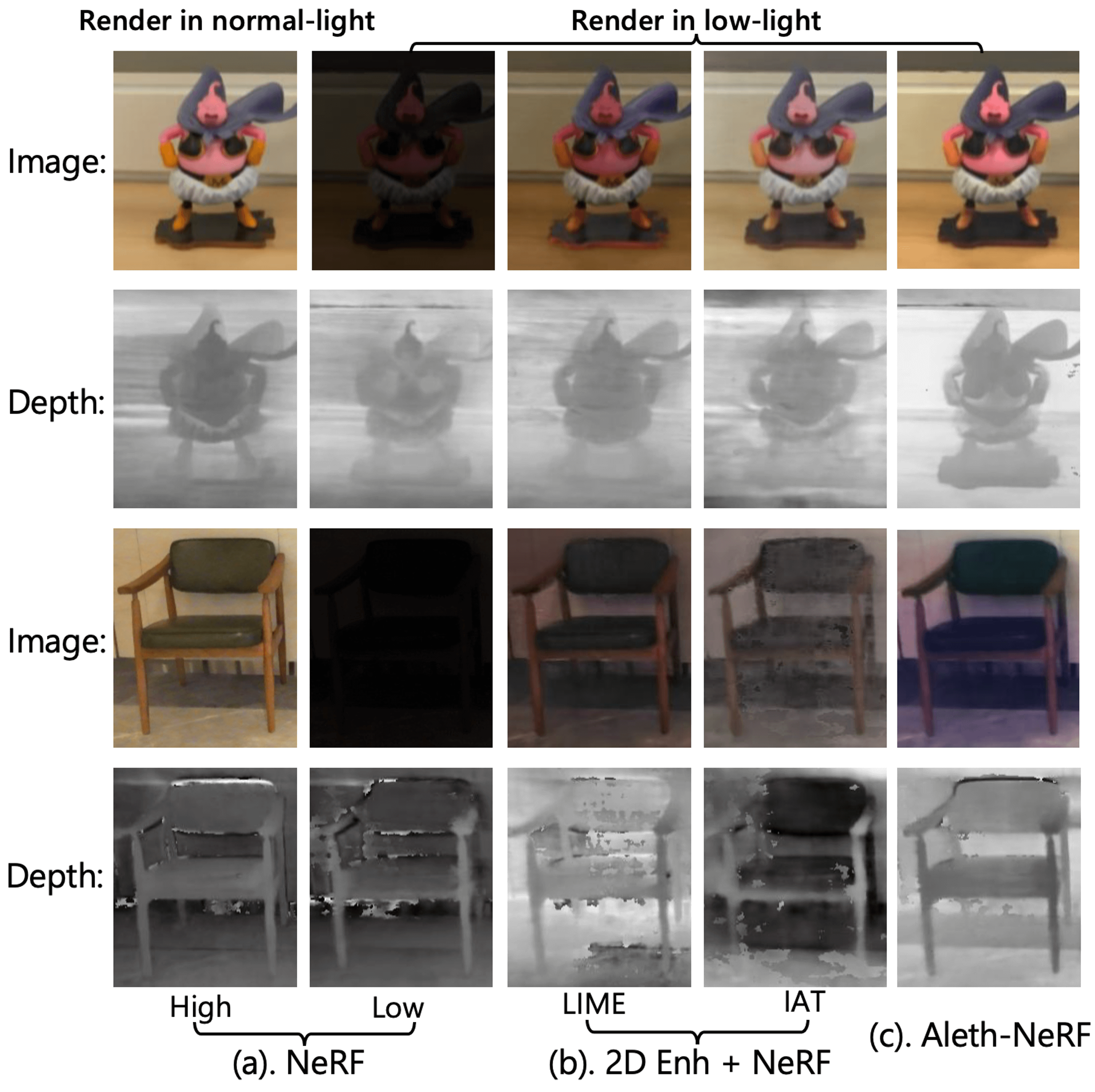}
    \caption{(a). NeRF rendering results in normal-light scene and low-light scene. (b). NeRF rendering on enhanced scene by 2D image enhancement methods LIME~\cite{LIME} and IAT~\cite{BMVC22_IAT}. (c) Aleth-NeRF rendering results in low-light scene.}
    \label{pics:motivation}
    \vspace{-3mm}
\end{figure}

This is because vanilla NeRF is \textit{viewer-centred} which  models the amount of light emission from a location to the viewer without counting the interaction between illumination and scenes. The emitted light results from the reflection of the light in the environment on the scene. The reflected light will be further refracted, absorbed, and thus attenuated in the environment again~\cite{srinivasan2021nerv}. 
As a result, the NeRF algorithm interprets a dark scene as resulting from insufficient radiation from the 3D particles representing objects in a scene. When training NeRF on dark images with a relatively high zero-mean noise signal~\cite{wei2020physics}, the algorithm may struggle to maintain multi-view projection consistency, leading to poor reconstruction results, shown in Fig.~\ref{pics:motivation}(a).


One solution is
extending NeRF to \textit{object-centred} rendering. That is, the observed lightness of the image is not due to the radiation intensity of the 3D particles representing the object, but rather to the attenuation of the light caused by other physical factors in the environment. However, this solution~\cite{srinivasan2021nerv,NeRFactor} usually requires known target lighting conditions and additional parametric modeling of the lighting.  

Leveraging mature 2D low-light image enhancement (LLIE) methods appear to be another solution. However, Fig.~\ref{pics:motivation}(b) shows that directly enhancing the dark images by 2D enhancement methods does not guarantee accurate NeRF estimation, since the independent and inconsistent enhancement of 2D images in multi-views may lead to the destruction of 3D geometric consistency.

To explore the NeRF application in these commonly captured but often precluded low-light scenes, we aim to propose a framework that directly supervises on low-light sRGB images. The rendering process in NeRF is similar to the emission theory held by ancient Greek. Emission theory ignores the incident light but postulates visual rays emitted from the eye travel in straight lines and interacts with objects to form the visual perception. Therefore, the darkness of an entity is solely caused by the particles between the object and the eye. In other words, all objects are visible by default unless they are concealed.  Inspired by this worldview, we assume a simple but NeRF-friendly model that it is the concealing fields in viewing direction makes the viewer see a low-light scene $\textbf{C}^{low}$ shown in Fig.\ref{pics:fig1}. Aletheia (\textalpha \textlambda\texteta\texttheta\textepsilon\textiota\textalpha), normally translated as "unconcealedness", "disclosure" or "revealing"~\cite{heidegger2010being}, removes the concealing fields to let right side viewer catch sight of normal-light scene $\hat{\textbf{C}}^{nor}$.

Here, we slightly modify the NeRF and propose our Aleth-NeRF for low-light conditions, which takes multi-view dark images as input to train the model and learn the volumetric representation jointly with Concealing Fields. As shown in Fig.\ref{pics:fig1}, we model this Concealing Field between object and viewer to naturally extend the transmittance function in NeRF~\cite{nerf}. In training stage, we aim to estimate not only the Radiance Fields representing the consistent objects, but also the Concealing Fields explaining the image transition from normal-light to low-light. In the testing stage, we could directly render the normal-lit scenes by simply eliminating Concealing Field. In addition, We collect the first paired low-light and normal-light multi-view dataset, \textbf{LOM}, to facilitate the understanding of 3D scenes under low-light conditions.

Our contribution could be summarized as follow: 

\begin{itemize}
    \item We propose \textbf{Aleth-NeRF}, by our knowledge, the first NeRF trains on dark multi-view sRGB images for unsupervised enhancement. Inspired by ancient Greek philosophy, we naturally extend the transmittance function in vanilla NeRF by modeling Concealing Fields between scene and viewer to interpret low light condition, endowing robustness to darkness with minimal modification on vanilla NeRF.
    \item We contribute the first low-light $\&$ normal-light paired multi-view real-world dataset \textbf{LOM}, along with comparisons with various 2D image enhancement methods, experiments show that our 
    \textbf{Aleth-NeRF} achieves satisfactory performance in both enhancement quality and multi-view consistency. 
    
\end{itemize}

\section{Related Work}
\label{sec:related_work}
\subsection{Low-light Image Enhancement}

Low-light image enhancement (LLIE) is a classical image processing task aiming to recover an image taken under inadequate illumination to its counterpart taken under normal illumination. Traditional LLIE methods usually make pre-defined illumination assumptions and use hand-crafted features for image restoration. For example, the RetiNex-based methods~\cite{retinex,multi_RetiNex,LIME} assume that a low-light image is multiplied by illumination and reflection. Histogram Equalization (HE) based methods~\cite{DIP_citation,Global_HE, Dynamic_HE} perform LLIE by spreading out the most frequent intensity values.


With the fast growth of deep learning in recent years, deep neural networks (DNNs) based methods have become the mainstream solutions these years. A series of CNN or Transformer-based methods have been developed in this area~\cite{Lv2018MBLLEN,RetiNexNet,Deep_LPF,LLFlow,Wu_2022_CVPR,Chen_2018_CVPR,Zheng_2021_ICCV,zhou2022lednet,zhang2022deep,RealTimeDarkImageRestorationCvpr2021,song2021starenhancer,BMVC22_IAT,LLFormer,wang2022lcdp}, which trained on paired low-light and normal-light images for supervision and gain satisfactory results. 

Beyond supervised methods, unsupervised low-light image enhancement is a more general but challenging setup, which does not require ground truth normal-light images for training. Some works like~\cite{Enlightengan,ECCV22_jin2022unsupervised} leverage the statistical illumination distribution in the target image domain for unsupervised training. Other works complete this task with only low-light information~\cite{Zero-DCE,SCI_CVPR2022,Kar_2021_CVPR,zhang2021_mm,RRDNet,wacv2023psenet}.
Like Zero-DCE~\cite{Zero-DCE} propose to train the network only with low-light images and regularize the network with modeled curves and some additional non-reference constraints (\textit{i.e.} color constancy~\cite{gray_world}). SCI~\cite{SCI_CVPR2022} completes this task with a self-calibrated module and unsupervised loss. 

However, current LLIE methods almost build on 2D image space operations, which often fail to exploit the 3D geometry of the scene and make it hard to deal with multi-view inputs.
The proposed Aleth-NeRF belongs to the unsupervised family but is empowered with a much stronger and more reliable understanding of 3D space. We follow the volume rendering formulation in NeRF and trace this problem back to the phase of image rendering from 3D radiant particles, where additional light concealing fields are introduced and learned unsupervisely for the LLIE problem, meanwhile maintaining the multi-view consistency.

\begin{figure*}
    \centering
    \includegraphics[width=0.95\linewidth]{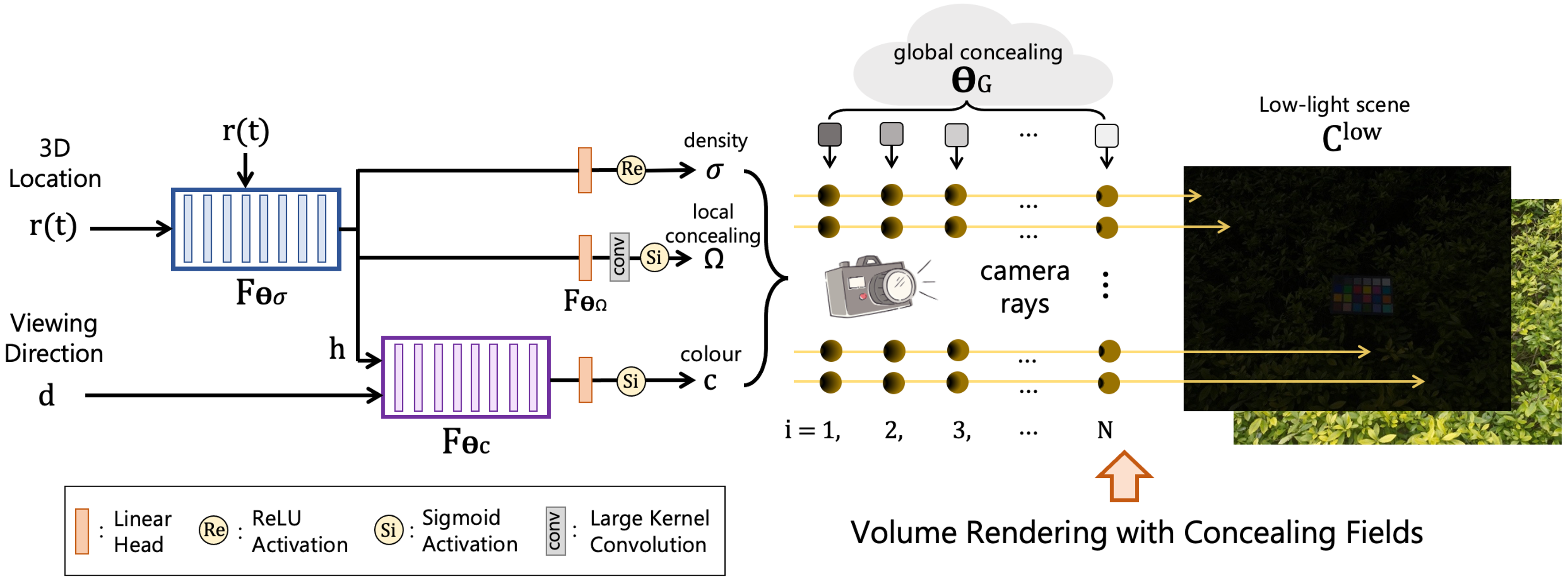}
    \caption{Overview of the Aleth-NeRF architecture. Local $\Omega$ and Global $\Theta_G$ Concealing Fields are additionally learned and integrated into the NeRF framework. We use a modified volume rendering function to render low-light scene taking the Concealing Fields into account.}
    \label{pics:fig2}
\end{figure*}

\subsection{Novel View synthesis with NeRF}

NeRF~\cite{nerf} is proposed for novel view synthesis from a collection of posed input images. It gains substantial attention even though the quality of the rendered image is yet to be comparable to state-of-the-art GAN~\cite{Style_gan_v3} or diffusion-based~\cite{DDPM,DDPM_Beat} models. The unique advantage of NeRF models exists in preserving the 3D geometry consistency thanks to its physical volume rendering scheme.

In addition to general efforts to speed up and improve NeRF training~\cite{barron2021mipnerf,yu_and_fridovichkeil2021plenoxels,autoint2021,yu2021plenoctrees,NeRF_on_DieT,roessle2022depthpriorsnerf,depth_nerf,mueller2022instant} or to use NeRF for scene understanding applications~\cite{semantic_Nerf,NARF_2021_ICCV,pose_inerf,Xu_2022_SinNeRF,NeRF-SOS}. Many of the latter works focus on improving NeRF's performance under various degradation conditions, such as blurry~\cite{deblur_nerf}, noisy~\cite{Nan_2022_CVPR}, reflection~\cite{NERF_reflection}, glossy surfaces~\cite{refnerf_CVPR22}, or use NeRF to complete low-level tasks in 3D space, like super-resolution~\cite{wang2021nerf-sr,volume_sr} and HDR reconstruction~\cite{hdrnerf,jun2022hdr}.


Another line of research extends NeRF for lightness editing in 3D space. Some work, like NeRF-W~\cite{nerf_wild}, focuses on rendering NeRF with uncontrolled in-the-wild images, other relighting works~\cite{srinivasan2021nerv,rudnev2022nerfosr,NeRFactor} rely on known illumination conditions and introduce additional physical elements (\textit{i.e.} normal, light, albedo, etc.), along with complex parametric modeling of these elements.
Meanwhile, these methods are not specifically designed for low-light scene enhancement under extreme low-light conditions.

Among these, RAW-NeRF~\cite{raw_nerf} is most close to our work, which proposes to render a NeRF model in nighttime RAW domain with high-dynamic range, and then post-process the rendered scene with image signal processor (ISP)~\cite{karaimer_brown_ECCV_2016}, RAW-NeRF has shown a preliminary ability to enhance the scene light but requires sufficient RAW data for training. We present the first work to render NeRF with low-light sRGB inputs and injection unsupervised LLIE into 3D space by an effective concealing fields manner.




\section{Method}

At first we briefly review core concepts of  vanilla NeRF on the radiance field, volume rendering, and image reconstruction loss in Section.~\ref{sec:nerf}. Then we introduce the proposed Concealing Fields in Section.~\ref{sec:night_nerf} and show how it is integrated into the NeRF framework to mimic the process of low-light scene generation. Finally, in Section.~\ref{sec.priors}, we show how to learn Concealing Field effectively in an unsupervised manner. 


\subsection{Neural Radiance Field Revisited}
\label{sec:nerf}
In the NeRF representation~\cite{nerf}, a Radiance Field is defined as the density $\sigma$ and RGB color value $c$ of a 3D location $\textbf{x}$ under a 2D viewing direction $\textbf{d}$. The density $\sigma$, on the one hand, represents the radiation capacity of the particle itself at $\textbf{x}$, and on the other hand, controls how much radiance is absorbed when other lights pass through $\textbf{x}$.

When rendering an image with a neural radiance field, a camera ray $\textbf{r}(t) = \textbf{o} + t \cdot \textbf{d}$ ($\textbf{r} \in \textbf{R}$), casting from the given camera position $\textbf{o}$ towards direction $\textbf{d}$, is used to accumulate all the radiance along the ray to render its corresponding pixel value $\textbf{C}(\textbf{r})$. This process is commonly implemented with the volume rendering function~\cite{volume_rendering}. Formally, 
\begin{equation}
{\bf C}({\bf r}) = \int_{t_n}^{t_f} T(\textbf{r}(t))\sigma({\bf r}(t))c({\bf r}(t), {\bf d})dt,
\end{equation}
 where
\begin{equation}
T(\textbf{r}(t)) = \exp (- \int_{t_n}^{t} \sigma({\bf r}(s))ds),    
\end{equation}
is known as the \emph{accumulated transmittance}. It denotes the radiance decay rate of the particle at $\textbf{r}(t)$ when it is occluded by particles closer to the camera (at $\textbf{r}(s)$, $s < t$).
The integrals are computed by a discrete approximation over sampled 3D points along the ray $\bf r$.
\begin{equation}
{\bf C}({\bf r}) = \sum_{i=1}^N T(\textbf{r}(i))(1 - \exp(-\sigma(\textbf{r}(i)) \cdot \delta)) \cdot c(\textbf{r}(i), \textbf{d})
\end{equation}
\begin{equation}
\label{eq.transmittance}
T(\textbf{r}(i)) = \exp \left(-\sum_{j=1}^{i-1}\sigma(\textbf{r}(j)) \cdot \delta \right) 
\end{equation}
where $\delta$ is a constant distance value between adjacent sample points under uniform sampling.

\paragraph{Neural Network Implementation.} NeRF learns two multilayer perceptron (MLP) networks ($F_{\Theta_{\sigma}}$, $F_{\Theta_{c}}$) to map the 3D location $\textbf{r}(i)$ and 2D viewing direction $\textbf{d}$ to its density $\sigma$ and colour $c$. Specifically,
\begin{equation}
F_{\Theta_{\sigma}}(\textbf{r}(i)) \rightarrow \sigma(\textbf{r}(i)), \textbf{h}
\end{equation}
\begin{equation}
F_{\Theta_{c}}(\textbf{h}, \textbf{d}) \rightarrow c(\textbf{r}(i), \textbf{d})
\end{equation}
where \textbf{h} is a hidden feature vector, $\Theta_{\sigma}$ and $\Theta_{c}$ are learnable network parameters. Note that $c(\textbf{r}(i), \textbf{d})$ and $\sigma(\textbf{r}(i))$ are further activated by Sigmoid and ReLU functions so that their value ranges are $[0, 1)$ and $[0, \infty)$, respectively.

Given ground truth rendered image $\textbf{C}$, the network is optimized by minimizing the image reconstruction loss between predicted image $\hat{\textbf{C}}$ and ground truth image $\textbf{C}$:

\begin{equation}
    \mathcal L_{nerf} = \sum_{\textbf{r}}^{\textbf{R}} || \hat{\textbf{C}}(\textbf{r}) - \textbf{C}(\textbf{r}) ||_2 .
\end{equation}
We refer techniques such as positional encoding and hierarchical volume sampling  to original paper~\cite{nerf} for more details. 

\begin{figure}
    \centering
    \includegraphics[width=1.00\linewidth]{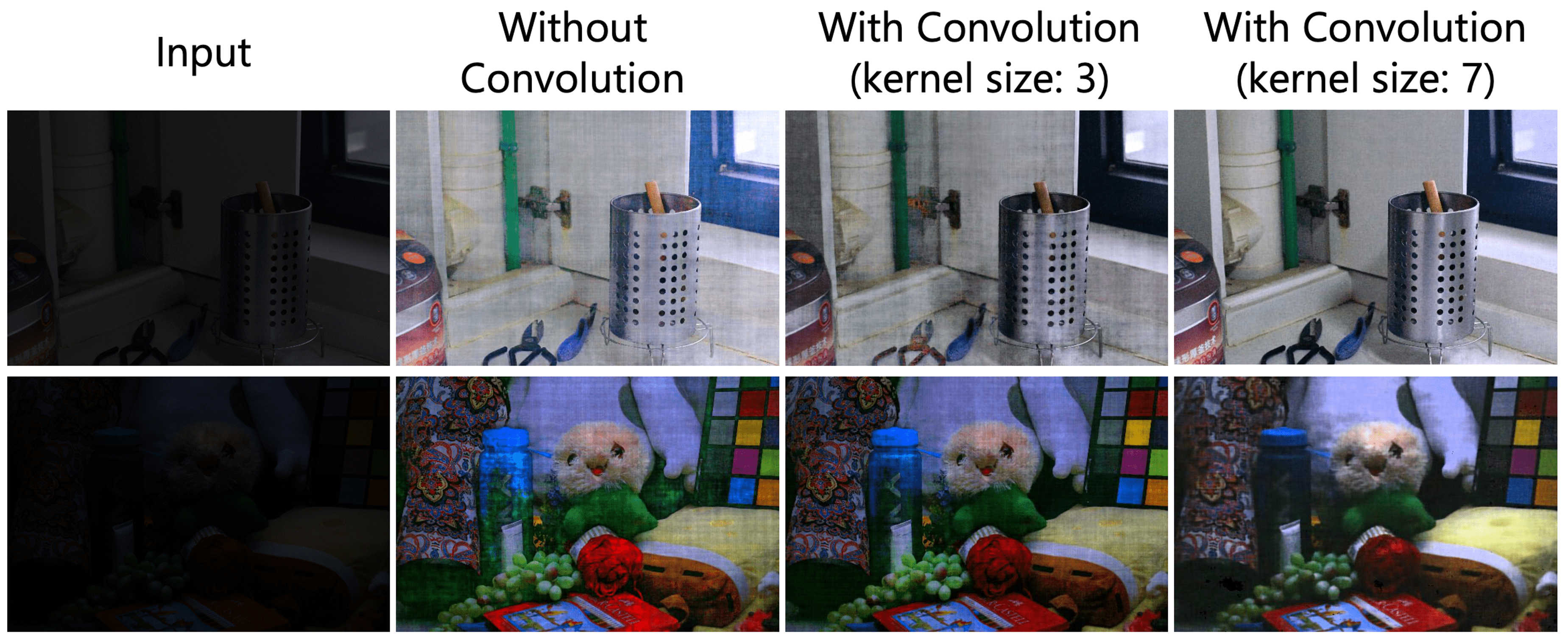}
    \vspace{-3mm}
    \caption{Ablation analyze on the LOL dataset~\cite{RetiNexNet}, larger convolution size to generate local concealing field $\Omega$ would further improve enhanced scene $\hat{\textbf{C}}^{nor}$'s smoothness.}
    \label{fig:conv_smooth}
\end{figure}

\subsection{Aleth-NeRF with Concealing Field Assumption}
\label{sec:night_nerf}
Given low-light scene $\{\textbf{C}^{low}(\textbf{r}), \textbf{r} \in \textbf{R}\}$ taken under poor illumination,
the goal of LLIE is to recover its normal-lit correspondence $\{\textbf{C}^{nor}(\textbf{r}), \textbf{r} \in \textbf{R}\}$. 
The key idea of our model is that we assume $\textbf{C}^{low}$ and $\textbf{C}^{nor}$ are rendered under the same Radiance Field condition (namely, they share the same underlying densities $\sigma$ and colors $c$ at all 3D locations in the scene) but with or without the proposed Concealing Field assumption.

\paragraph{Global and Local Concealing Fields}
We design two types of Concealing Fields, namely the local Concealing Field $\Omega$ and global Concealing Field $\Theta_G$ for low-light scene generation. Specifically, $\Omega$ controls the light concealing at voxel level while $\Theta_G$ controls at scene level.

The local Concealing Field, denoted as $\Omega(\textbf{r}(i))$, defines an extra light concealing capacity of a particle at 3D location $\textbf{r}(i)$. As it shown in Fig.~\ref{pics:fig2}, $\Omega(\textbf{r}(i))$ is learned for each 3D location,
for implementation we add a linear layer head  $F_{\Theta_{\Omega}}$ upon the first MLP network $F_{\Theta_{\sigma}}$, an additional convolution layer has been added after $F_{\Theta_{\Omega}}$ to generate the local concealing $\Omega(\textbf{r}(i))$, convolution process could build spatial relations between pixels and let Concealing Field contain more light information rather than structure information~\cite{Zamir2020CycleISP}, make rendering results smooth (see Fig.~\ref{fig:conv_smooth}).

\begin{equation}
F_{\Theta_{\Omega}}(F_{\Theta_{\sigma}}(\textbf{r}(i))) \rightarrow \Omega(\textbf{r}(i))
\end{equation}

The global-wise Concealing Field, denoted as $\Theta_G(i)$, is defined as a set of learnable parameters corresponding to the camera distance $i$ for all camera rays in $\textbf{R}$. $\Theta_G(i)$ is kept the same in a scene rendering and irrelevant to pixel level.



In the training stage, when we render and reconstruct a given low-light image $\textbf{C}^{low}$ with the reconstruction loss $\mathcal L_{nerf}$ computed between predicted $\hat{\textbf{C}}^{low}$ and ground truth $\textbf{C}^{low}$ images, the \emph{accumulated transmittance} $T$ in Eq.~\ref{eq.transmittance} is further modulated with the additional local Concealing Field $\Omega$ and global Concealing Field $\Theta_G$, to mimic the process of light suppression in a normal scene:

\begin{equation}
\label{eq.transmittance_local_global}
T^{low}(\textbf{r}(i)) = \exp \left(-\sum_{j=1}^{i-1}\sigma(\textbf{r}(j)) \cdot \delta \right) \cdot \prod_{j=1}^{i-1} \Omega(\textbf{r}(j)) \Theta_G(j)
\end{equation}

The testing stage performs aletheia or unconcealedness to remove Concealing Fields $\Omega$ and $\Theta_G$  to render the predicted normal-light image $\hat{\textbf{C}}^{nor}$ directly.
As a result, we transform the unsupervised LLIE problem into unsupervised learning of  Concealing Fields in the low-light scene. By adding or removing the Concealing Fields, Aleth-NeRF can render images under low-light or normal-light conditions of the same scene easily.


\begin{figure}
    \centering
    \includegraphics[width=1.00\linewidth]{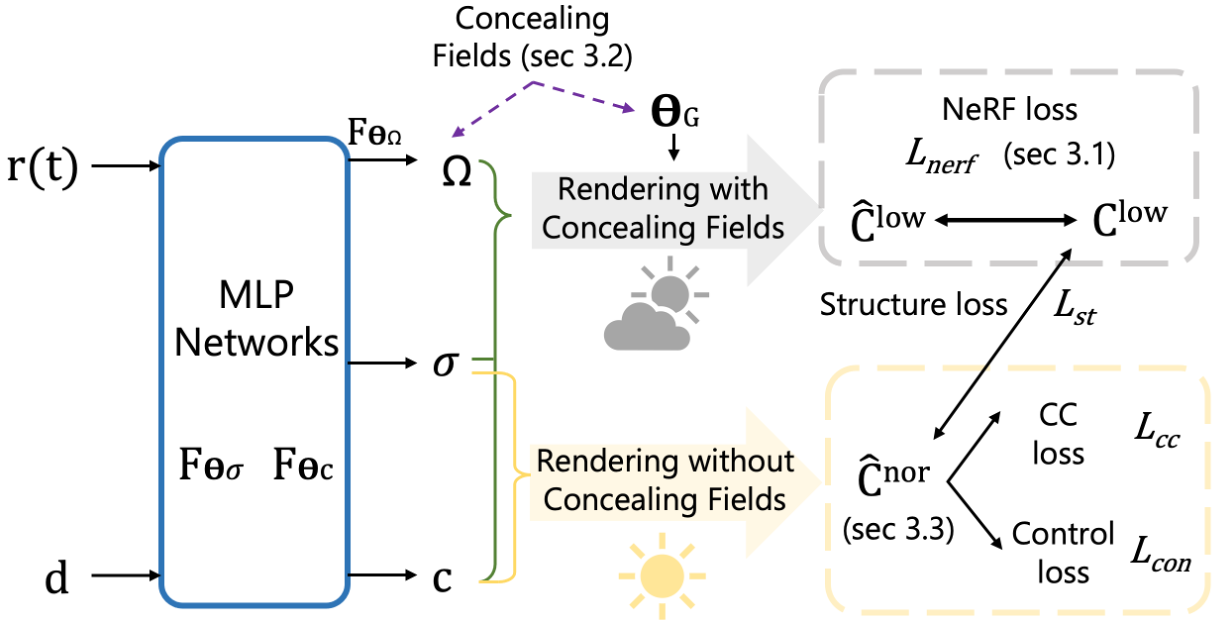}
    \caption{Aleth-NeRF uses reconstruction loss ($\mathcal L_{nerf}$) on the low-light scene $\textbf{C}^{low}$, meanwhile additional constraints (control loss $\mathcal L_{con}$, structure loss $\mathcal L_{st}$ and color constancy loss $\mathcal L_{cc}$) are added to regularize the predicted normal-light scene $\hat{\textbf{C}}^{nor}$.}
    \label{fig:loss}
\end{figure}

\subsection{Effective Priors for Unsupervised Training}
\label{sec.priors}
In this section, we introduce the priors and constraints that promote unsupervised learning of Concealing Fields. Training strategy overview is shown in Fig.~\ref{fig:loss}.

\textbf{Value Range Prior.}
Value range of concealing fields would determine the enhancement scene $\hat{\textbf{C}}^{nor}$'s lightness, generally the concealing fields' range should be less than $1$, for local concealing field $\Omega$, we add an $sigmoid$ activation after the convolution process (see Fig.~\ref{pics:fig2}), the activated $\Omega$ would lie in the range of $\left(0, 1 \right)$. Meanwhile, the global concealing field $\Theta_G(i)$ is a set of learnable parameters in the network, the initial value of $\Theta_G(i)$ is all set to $0.3$ along the camera ray. 


To control the degree of image enhancement, we add a control loss $\mathcal L_{con}$ on the local concealing field $\Omega$, where we introduce a hyper-parameter: $\eta$, representing the degree of Aleth-NeRF's concealing ability. To calculate $\mathcal L_{con}$, we apply an average pooling (stride $64$) to local concealing field $\Omega$, then minimize loss $\mathcal L_{con}$ between the pooled $\Omega$'s mean and the concealing degree $\eta$:
\begin{equation}
    \mathcal L_{con} = || \rm{avgpool}(\Omega(\textbf{r}(i))) - \eta ||^2,
\end{equation}
 Here, the concealing degree $\eta$ should be set larger than 0. In our work, we set $\eta$ to $0.05$ in extremely dark low-light conditions and to $0.1$ in conditions not so dark. 

\textbf{Structure Similarity Prior.} 
To maintain the structure information of the predicted normal light images $\hat{\textbf{C}}^{nor}$ consistent with the original low light image $\textbf{C}^{low}$, we additionally add a structure loss between them, to maintain the detail information and increase the contrast. Specifically, for each rendered pixel $\hat{\textbf{C}}^{nor} (\textbf{r})$, we keep a structure consistency with neighbored pixel $\hat{\textbf{C}}^{nor} (\textbf{r}-1)$ and $\hat{\textbf{C}}^{nor} (\textbf{r}+1)$ along the ray as follows:

\begin{equation}
\begin{aligned}
     \mathcal L_{st} = & \sum_{k \in [+1, -1]} (\hat{\textbf{C}}^{nor} (\textbf{r}) - \hat{\textbf{C}}^{nor} (\textbf{r}+k)) - \\ & \frac{0.5}{\eta} (\textbf{C}^{low} (\textbf{r}) - \textbf{C}^{low} (\textbf{r}+k)),
\end{aligned}
\end{equation}
here $\frac{0.5}{\eta}$ determines generated scene $\hat{\textbf{C}}^{nor}$'s contrast degree, which is inversely proportional to conceal degree $\eta$. More ablation analyses on conceal degree and contrast degree are in our supplementary.

\textbf{Color Constancy Prior.}  
Pixels taken under low-light conditions would lose some color information, and direct take off the concealing fields would easily cause color imbalance. To regularize the color of the predicted normal-light images $\hat{\textbf{C}}^{nor}$, we add an extra color constancy loss $\mathcal L_{cc}$ on $\hat{\textbf{C}}^{nor}$. Here we assume that $\hat{\textbf{C}}^{nor}$ obey the gray-world assumption~\cite{gray_world,Zero-DCE,ECCV22_jin2022unsupervised}, as follows:

\begin{equation}
    \mathcal L_{cc} = \sum_{p, q} (\hat{\textbf{C}}^{nor} (\textbf{r})^p - \hat{\textbf{C}}^{nor} (\textbf{r})^q)^2,
\end{equation}
where $(p, q) \in \{ (R, G), (G, B), (B, R) \}$ represents any pair of color channels.

Above all, the total loss function $\mathcal L_{total}$ of Aleth-NeRF include 4 parts, low-light scene rendering loss $\mathcal L_{nerf}$ and additional constrain losses $\mathcal L_{con}$, $\mathcal L_{st}$, $\mathcal L_{cc}$, the total loss used for training can be represented as: 
\begin{equation}
    \mathcal L_{total} = \mathcal L_{nerf} + \lambda_1 \cdot \mathcal L_{con} + \lambda_2 \cdot \mathcal L_{st} + \lambda_3 \cdot \mathcal L_{cc},
\end{equation}
where $\lambda_1$, $\lambda_2$ and $\lambda_3$ are three non-negative parameters to balance total loss weights, which we set to $1e^{-4}$, $1e^{-3}$ and $1e^{-4}$ respectively.

\begin{figure*}[t]
    \centering
    \includegraphics[width=1.00\linewidth]{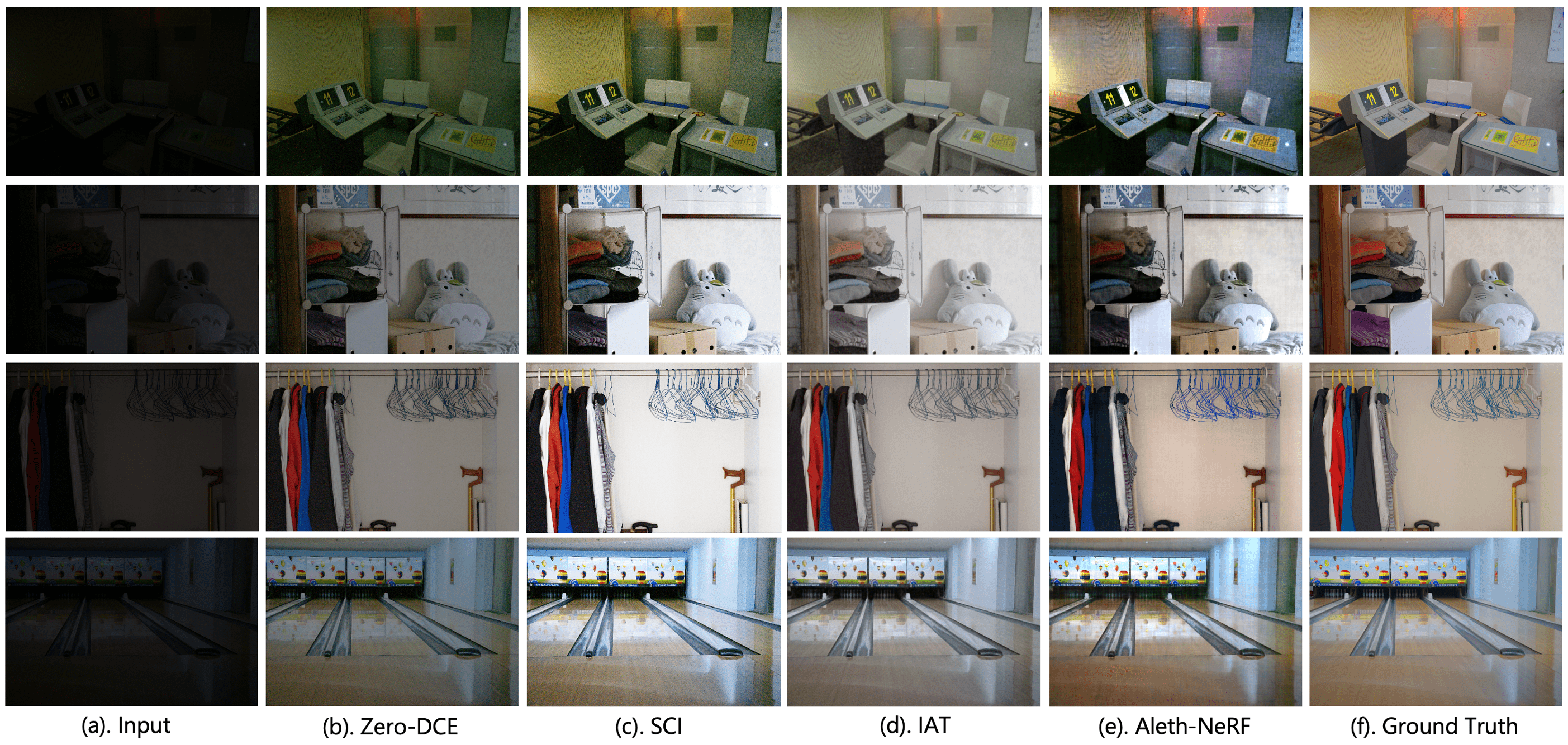}
    \vspace{-3mm}
    \caption{Example of enhancement results on LOL~\cite{RetiNexNet} evaluation set, (a) is input image, then are the enhancements results by (b). Zero-DCE~\cite{Zero-DCE}, (c). SCI~\cite{SCI_CVPR2022}, (d). IAT~\cite{BMVC22_IAT} and (e) our \textbf{Aleth-NeRF}, (f) is the ground truth image.}
    \label{fig:LOL_results}
\end{figure*}

\paragraph{Discussion} Although our method was originally designed for multi-view conditions, surprisingly, it achieves the same excellent low-light enhancement result for single-view images (see experiments in Sec.~\ref{exp:single_image}). Despite the ambiguity of depth estimation and the lack of synthesis capability for novel views, we believe that the success of single-view low-light enhancement greatly benefits from the effective priors we present in this section.

\section{Experiments}

In this section, we first benchmark on single image  low-light enhancement dataset LOL~\cite{RetiNexNet} in Sec.~\ref{exp:single_image}. Then we would introduce our collected \textbf{LOM} dataset with paired multi-view low-light and normal-light images in Sec.~\ref{exp:dataset}. Multi-view experimental results and analyses are shown in Sec.~\ref{exp:multi_view}.


Our framework is base on \texttt{NeRF-Factory}~\cite{nerf_factory} toolbox. We adopt the Adam optimizer with the initial learning rate set to $5e^{-4}$. Besides cosine learning rate decay strategy are set at every 2500 iters.
For the single image experiments on LOL~\cite{RetiNexNet} dataset, the camera position $\textbf{o}$ and viewing direction $\textbf{d}$ are fixed, and batch size is set to 8192 for 20000 iters training. For multi-view experiments on \textbf{LOM} dataset, batch size is set to 4096 for 62500 iters training. 

\subsection{Generation Quality Assessment}
\label{exp:single_image}

\begin{table}[t]
\renewcommand\arraystretch{1.33}
\caption{Quantitative comparison on LOL~\cite{RetiNexNet} dataset, here * denotes the zero-shot methods.}
\vspace{1.5mm}
\label{table:LOL_results}
\centering
\begin{adjustbox}{max width = 0.99\linewidth}
\begin{tabular}{cc|ccc}

\Xhline{1.0pt}
\multicolumn{2}{c|}{Method}                                       &  PSNR $\uparrow$ & SSIM $\uparrow$ & LPIPS $\downarrow$  \\ \Xhline{1.0pt}
\multicolumn{1}{c|}{\multirow{3}{*}{supervised}}   & RetiNexNet~\cite{RetiNexNet}   & 16.67 & 0.562 & 0.474 \\ \cline{2-5} 
\multicolumn{1}{c|}{}                              & MBLLEN~\cite{Lv2018MBLLEN}       & 17.96 & 0.756 & 0.386 \\ \cline{2-5} 
\multicolumn{1}{c|}{}                              & KIND~\cite{KIND}        &      20.86 &     0.790   &  0.448     \\ \hline

\multicolumn{1}{c|}{\multirow{7}{*}{unsupervised}} &

HE*~\cite{DIP_citation}         & 14.54 & 0.376 & 0.466 \\ \cline{2-5} 
\multicolumn{1}{c|}{}   &

LIME*~\cite{LIME}         & 14.01 & 0.513 & 0.391 \\ \cline{2-5} 
\multicolumn{1}{c|}{}   & 
Zero-DCE~\cite{Zero-DCE}     & 14.63 & 0.558 & \textbf{0.385} \\ \cline{2-5} 
                           
\multicolumn{1}{c|}{}     & Zero-Restore*~\cite{Kar_2021_CVPR} &   14.03    &  0.460  & 0.451 \\ \cline{2-5} 

\multicolumn{1}{c|}{}          & RRDNet*~\cite{RRDNet} &    11.39   &   0.468     &   0.451    \\ \cline{2-5} 

\multicolumn{1}{c|}{}                              
& En-GAN~\cite{Enlightengan}      &  12.23    &  0.596    &    0.498   \\ \cline{2-5}

\multicolumn{1}{c|}{}                            & SCI~\cite{SCI_CVPR2022}     & 15.65 & 0.510 & 0.419 \\ \cline{2-5}  

\multicolumn{1}{c|}{}                              & \textbf{Aleth-NeRF}*         &   \textbf{17.14}    &   \textbf{0.653}    &  0.445     \\ \Xhline{1.0pt}
\end{tabular}
\end{adjustbox}

\end{table}

We first conduct the experiments on benchmark single image low-light enhancement dataset LOL~\cite{RetiNexNet}. LOL consists of 500 paired low-light images $\textbf{C}^{low}$ and normal-light images $\textbf{C}^{nor}$, 485 pairs are split into a train set, and the other 15 pairs are split into evaluation set. 

During training stage, we only learn the 15 low-light images $\textbf{C}^{low}$ in the evaluation set. Without either the reference normal-light images $\textbf{C}^{nor}$ or the other 485 low-light images in the train set. The quantitative comparison with various low-light enhancement methods~\cite{DIP_citation,RetiNexNet,Lv2018MBLLEN,KIND,Zero-DCE,LIME,Enlightengan,Kar_2021_CVPR,SCI_CVPR2022,RRDNet} is shown in Table.~\ref{table:LOL_results}, we report three image quality metrics: PSNR, SSIM and LPIPS~\cite{lpips}. Table.~\ref{table:LOL_results} shows that our methods gain satisfactory results among the unsupervised methods, ensuring the \textbf{generation quality} of Aleth-NeRF. We show 4 examples in Fig.~\ref{fig:LOL_results}, including other enhancement results by current SOTA methods Zero-DCE~\cite{Zero-DCE}, SCI~\cite{SCI_CVPR2022} and IAT~\cite{BMVC22_IAT}. Fig.~\ref{fig:LOL_results} shows that our Aleth-NeRF could gain high-quality and vivid results compared with various 2D enhancement methods.

\begin{table*}[]
\renewcommand\arraystretch{1.33}
\caption{LOM~\cite{RetiNexNet} dataset results, we evaluate PSNR $\uparrow$ (higher the better), SSIM $\uparrow$ (higher the better) and LPIPS $\downarrow$ (lower the better).}
\label{table:LOM_results}
\centering
\begin{adjustbox}{max width = 1.00\linewidth}
\begin{tabular}{l|c|c|c|c|c|c}

\toprule
\toprule
\multirow{2}{*}{method} & ``\textbf{\textit{buu}}"              & ``\textbf{\textit{chair}}"              & ``\textbf{\textit{sofa}}"                & ``\textbf{\textit{bike}}"                & ``\textbf{\textit{shrub}}"               & \textbf{\textit{mean}}                \\ \cline{2-7} 
                        & PSNR/ SSIM/ LPIPS  & PSNR/ SSIM/ LPIPS   & PSNR/ SSIM/ LPIPS   & PSNR/ SSIM/ LPIPS   & PSNR/ SSIM/ LPIPS   & PSNR/ SSIM/ LPIPS   \\ \hline \hline
NeRF~\cite{nerf}                    & \enspace 7.53/ 0.313/ 0.414  & \enspace 6.01/ 0.142/ 0.599  & 6.26/ 0.206/ 0.566  & \enspace 6.32/ 0.067/ 0.625  & \enspace 8.00/ 0.027/ 0.684  & \enspace 5.24/ 0.151/ 0.578  \\
NeRF + HE~\cite{DIP_citation}               & 14.90/ 0.678/ 0.578 & 15.11/ 0.615/ 0.658 & 17.28/ 0.713/ 0.624 & 14.26/ 0.547/ 0.582 & 12.40/ 0.384/ 0.647 & 14.79/ 0.587/ 0.618 \\
NeRF + LIME~\cite{LIME}             & 13.67/ 0.745/ 0.375 & 11.11/ 0.637/ 0.600 & 12.28/ 0.723/ 0.562 & 10.82/ 0.504/ 0.557 & 14.12/ 0.360/ 0.540 & 12.40/ 0.593/ 0.526 \\
NeRF + RetiNexNet~\cite{RetiNexNet}       & 16.18/ 0.754/ 0.385 & 16.74/ 0.731/ 0.531 & 15.90/ 0.826/ 0.512 & 17.79/ 0.654/ 0.552 & 15.24/ 0.283/ 0.577 & 16.37/ 0.657/ 0.495 \\
NeRF + SCI~\cite{SCI_CVPR2022}              & 13.66/ 0.769/ 0.433 & 18.42/ 0.736/ 0.545 & 19.82/ 0.813/ 0.569 & 12.61/ 0.560/ 0.550 & 16.87/ 0.411/ 0.555 & 16.27/ 0.658/ 0.530 \\
NeRF + IAT~\cite{BMVC22_IAT}              & 14.00/ 0.652/ 0.421 & 19.53/ 0.806/ 0.562 & 10.58/ 0.532/ 0.668 & 16.30/ 0.660/ 0.532 & 10.03/ 0.149/ 0.577 & 14.09/ 0.560/ 0.552 \\ \hline \hline
HE~\cite{DIP_citation} + NeRF               & 15.54/ 0.755/ 0.473 & 15.46/ 0.756/ 0.492 & 17.89/ 0.801/ 0.457 & 15.47/ 0.699/ 0.455 & 14.97/ 0.414/ 0.511 & 15.87/ 0.685/ 0.470 \\
LIME~\cite{LIME} + NeRF             & 13.81/ 0.783/ 0.253 & 11.20/ 0.694/ 0.498 & 12.02/ 0.747/ 0.420 & 11.35/ 0.586/ 0.448 & 14.11/ 0.426/ \textbf{0.473} & 12.50/ 0.647/ 0.418 \\
RetiNexNet~\cite{RetiNexNet} + NeRF       & 16.16/ 0.777/ 0.339 & 16.82/ 0.773/ \textbf{0.438} & 16.87/ 0.806/ 0.548 & 18.00/ \textbf{0.717}/ 0.448 &    14.65/ 0.269/ 0.513     &    16.50/ 0.668/ 0.457   \\
SCI~\cite{SCI_CVPR2022} + NeRF              & 13.62/ 0.821/ 0.309 &    11.75/ 0.756/ 0.526  & 10.10/ 0.750/ 0.505 & \textbf{19.10}/ 0.644/ 0.458  &              18.13/ 0.510/ 0.469        &   14.54/ 0.696/ 0.453 \\
IAT~\cite{BMVC22_IAT} + NeRF              & 14.33/ 0.697/ 0.311 & 18.68/ 0.789/ 0.563 & 17.77/ 0.811/ 0.519 & 13.68/ 0.621/ 0.501 & 13.84/ 0.322/ 0.536 & 15.66/ 0.648/ 0.486 \\ \hline \hline
\textbf{Aleth-NeRF}              & \textbf{18.93}/ \textbf{0.825}/ \textbf{0.251} & \textbf{19.71}/ \textbf{0.812}/ 0.476 & \textbf{19.98}/ \textbf{0.851}/ \textbf{0.411} & 15.53/ 0.691/ \textbf{0.444} & \textbf{18.31}/ \textbf{0.511}/ 0.506 & \textbf{18.49}/ \textbf{0.738}/ \textbf{0.417} \\ \bottomrule \bottomrule
\end{tabular}
\end{adjustbox}
\end{table*}

\begin{figure}
    \centering
    \includegraphics[width=1.00\linewidth]{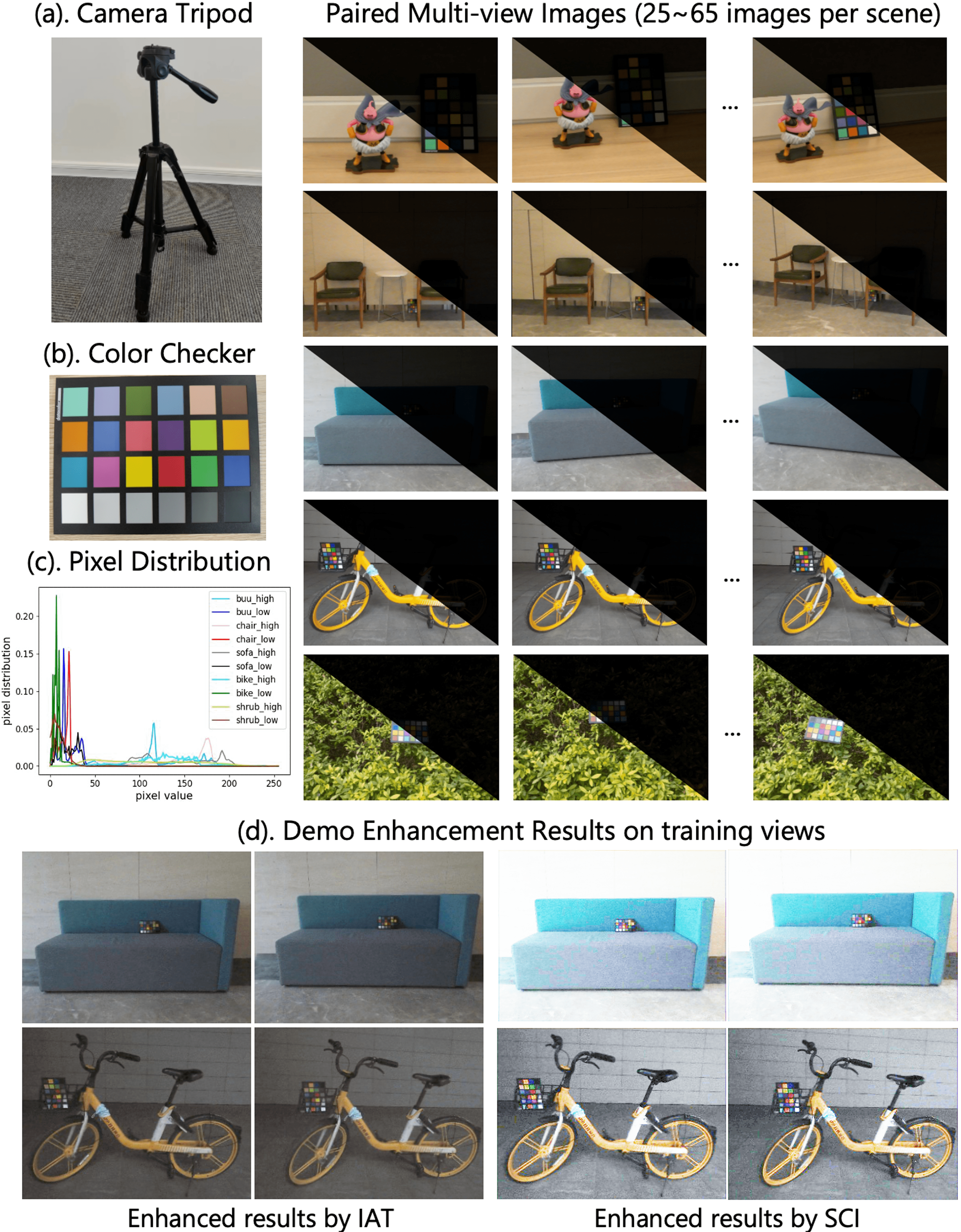}
    \caption{Collection detail of the \textbf{LOM} dataset.}
    \label{fig:LOM_collection}
\end{figure}

\subsection{Multi-view Paired Dataset in Real-world}
\label{exp:dataset}
In this section, we introduce our collected paired low-light and normal-light multi-view dataset, namely \textbf{LOM} dataset. To the best of our knowledge, this is the first dataset containing paired multi-view low-light and normal-light images. Previous work~\cite{raw_nerf} also includes some low-light scenes. However, their dataset concentrates on RAW denoising rather than sRGB low-light enhancement. Besides~\cite{raw_nerf} does not include normal-light sRGB counterpart, making it hard to evaluate the model performance of multi-view low-light enhancement in real-world. 

\textbf{LOM} dataset has 5 real-world scenes  (``\textbf{\textit{buu}}", ``\textbf{\textit{chair}}", ``\textbf{\textit{sofa}}", ``\textbf{\textit{bike}}", ``\textbf{\textit{shrub}}"). Each scene includes $25 \sim 65$ low-light and normal-light pairs. We collect the scenes with DJI Osmo Action 3 camera, then generate pair low-light and normal-light images by adjusting exposure time and ISO while other configurations of the camera are fixed. We also take a camera tripod to prevent camera shake (Fig.~\ref{fig:LOM_collection}(a)), we capture multi-view images by moving and rotating the tripod. Additionally, for each scene, we add a DSLR color checker (Fig.~\ref{fig:LOM_collection}(b)) to help us determine the color and better evaluate generated images. Images are collected with resolution 3000 $\times$ 4000. We down-sample the original resolution with ratio 8 to 375 $\times$ 500 for convenience, and generate the ground truth view and angle information by adopting COLMAP~\cite{COLMAP1,COLMAP2} on the normal-light scenes. For dataset split, in each scene, we choose 3 $\sim$ 5 images as the testing set, 1 image as the validation set, and other images to be the training set. The Y-channel pixel distribution of each scene has been shown in Fig.~\ref{fig:LOM_collection}(c). Exemplary low-light scene enhancement  results  by SOTA 2D enhancement methods IAT~\cite{BMVC22_IAT} and SCI~\cite{SCI_CVPR2022} are shown in Fig.~\ref{fig:LOM_collection}(d).

\begin{figure*}[t]
    \centering
    \includegraphics[width=1.00\linewidth]{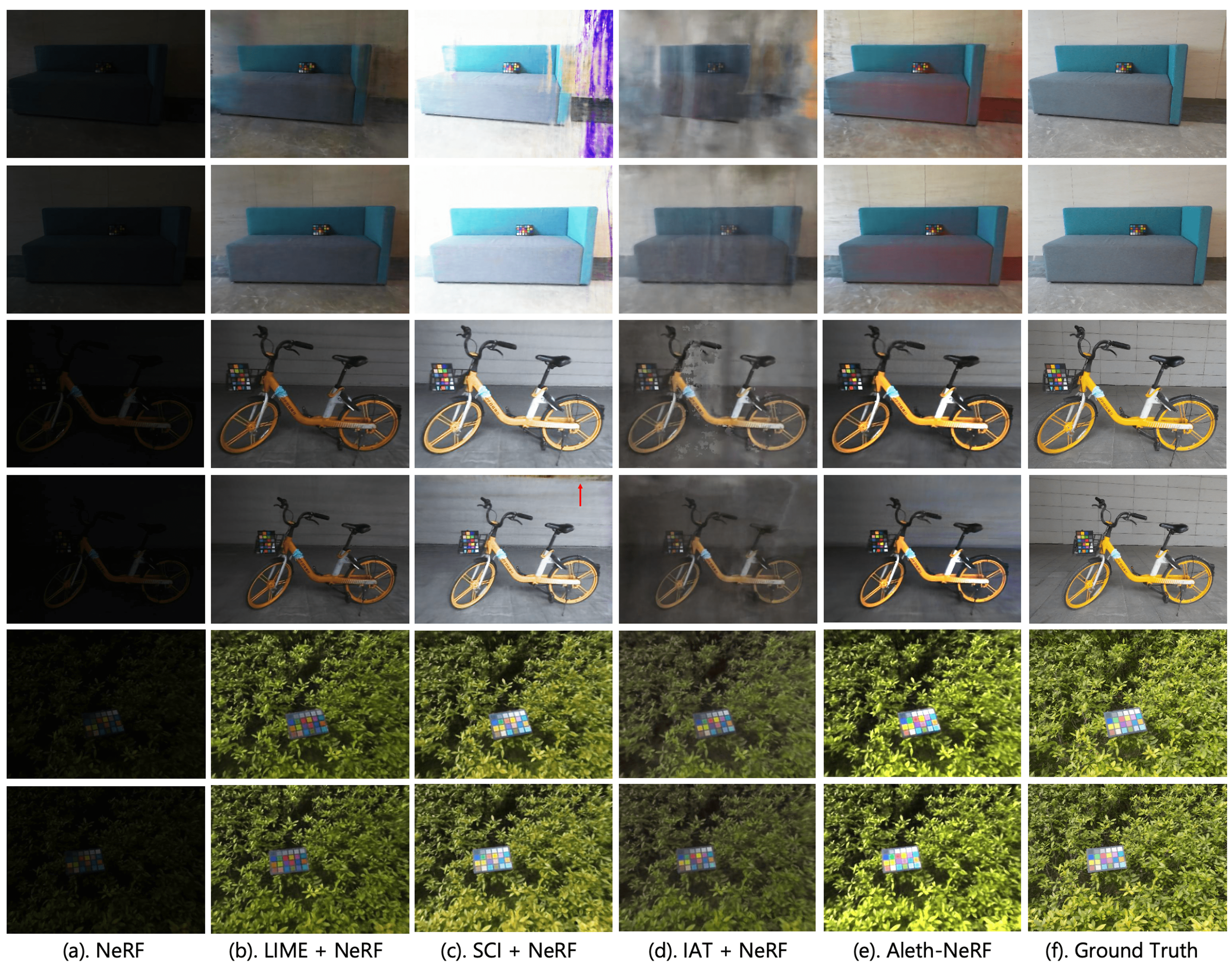}
    \caption{Multi-view rendering results on \textbf{LOM} dataset's ``\textbf{\textit{sofa}}", ``\textbf{\textit{bike}}" and ``\textbf{\textit{shrub}"} scenes.}
    \label{pics:LOM_results}
    \vspace{-2mm}
\end{figure*}

\subsection{Multi-view Enhancement Results}
\label{exp:multi_view}
In this scention, we show multi-view rendering results on low-light scenes in \textbf{LOM} dataset. We design multiple comparison experiments to evaluate the generation quality and multi-view consistency. We first make the comparison with the vanilla NeRF~\cite{nerf}, which has the same setting as Aleth-NeRF that only trains with low-light scene images $\textbf{C}^{low}$. Then we compare with five low-light enhancement methods: histogram equalization (HE)~\cite{DIP_citation}, LIME~\cite{LIME}, RetiNexNet~\cite{RetiNexNet}, IAT~\cite{BMVC22_IAT} and SCI~\cite{SCI_CVPR2022}, here HE and LIME are two traditional enhancement method, IAT and SCI are two very recent SOTA network-based 2D enhancement methods. As shown in Table~\ref{table:LOM_results}, we design two settings for comparison, (1). Rendering low-light scenes by NeRF and then using 2D enhancement methods to post-process these low-light novel views, name as ``NeRF + *". (2). Using 2D enhancement methods to pre-process training set and rendering NeRF on these enhanced views, name as ``* + NeRF". All the comparison experiments take the same training epochs, batch size, learning rate, and strategies to ensure fairness.

The comparison results are shown in Table.~\ref{table:LOM_results}. All rendering results would compare with LOM's normal-lit view counterpart. We report to three image quality metrics: SSIM, PSNR and LPIPS~\cite{lpips}. The results of each scene are then averaged to generate the \textbf{\textit{mean}}  results (last column in Table.~\ref{table:LOM_results}). From the results, we could find that ``NeRF + *" series methods almost perform worse than ``* + NeRF" series methods. This is probably due to the poor image quality NeRF produces in dark scenes, making enhancement methods easily fail on the generated dark scenes by NeRF. Meanwhile, for the ``* + NeRF" series methods, although the enhancement quality performs well on the training set (see Fig.~\ref{fig:LOM_collection}(d)), 2D enhancement methods lack multi-view consistency and often fail on novel view generation (see Fig.~\ref{pics:LOM_results}), make it easier to generate artifacts and noise. Overall, our Aleth-NeRF is an end-to-end method and gain the best performance on most scenes and the \textbf{\textit{mean}} results, the generated novel views both maintain multi-view consistency and image generation quality. Please refer to our supplementary for more visualization results and ablation analyze.


\section{Conclusion and Discussion}

We propose a novel unsupervised method to handle multi-view synthesis in low-light condition, which directly take low-light scene as input and render out normal-light scene. Inspired by the wisdom of the ancient Greeks, we introduce a concept: Concealing Fields. Experiments demonstrate our superior performance in both image quality and 3D multi-view consistency.


One limitation is that Aleth-NeRF should be specifically trained for each scene, which is the same as original NeRF~\cite{nerf}. Besides, Aleth-NeRF may fail in scenes with non-uniform lighting conditions or shadow conditions. We'll solve these problems in the future.

\section{Acknowledgments}
This work was partially supported by JST Moonshot R$\&$D Grant Number JPMJPS2011, CREST Grant Number JPMJCR2015 and Basic Research Grant (Super AI) of Institute for AI and Beyond of the University of Tokyo. Also
this work is partially supported by the National Key R$\&$D Program of China(NO.2022ZD0160100),and in part by Shanghai Committee of Science and Technology (Grant No. 21DZ1100100).


{\small
\bibliographystyle{ieee_fullname}
\bibliography{egbib}
}

\end{document}